\documentclass[conference]{IEEEtran}
\usepackage{tikz}
\usepackage{stmaryrd}
\usepackage{amsfonts}

\usepackage{graphicx,times,psfig,amsmath} 

\IEEEoverridecommandlockouts    

\begin{document}

\title{\ \\ \LARGE\bf Tuning a Multiple Classifier System for Side Effect Discovery using Genetic Algorithms \thanks{Jenna M. Reps, Uwe Aickelin and Jonathan M. Garibaldi are with the School of Computer Science, The University of Nottingham, UK (email: \{jenna.reps, uwe.aickelin, jonathan.garibaldi \}@nottingham.ac.uk).} \thanks{}}

\author{Jenna M. Reps, Uwe Aickelin and Jonathan M. Garibaldi}


\maketitle

\begin{abstract}
In previous work, a novel supervised framework implementing a binary classifier was presented that obtained excellent results for side effect discovery.  Interestingly, unique side effects were identified when different binary classifiers were used within the framework, prompting the investigation of applying a multiple classifier system.  In this paper we investigate tuning a side effect multiple classifying system using genetic algorithms.  The results of this research show that the novel framework implementing a multiple classifying system trained using genetic algorithms can obtain a higher partial area under the receiver operating characteristic curve than implementing a single classifier.  Furthermore, the framework is able to detect side effects efficiently and obtains a low false positive rate.
\end{abstract}


\section{Introduction}

\PARstart{S}{ide effects} of prescription drugs are a common occurrence that often lead to patient morbidity and mortality.  When there is an association between a medical event (e.g., sickness, rash and weakness) and a drug, this is termed an adverse event (AE).  When the relationship is proven to be causal (i.e., the drug causes the medical event), it is referred to as an adverse drug reaction (ADR). 

As a large quantity of medical data are often stored in databases, numerous methods have been presented that make use of medical databases with the aim of identifying ADRs efficiently \cite{Noren2010,Zorych2013}.  Unfortunately, the majority of these methods work by finding medical events that are highly associated to a drug, therefore, rather than detecting ADRs they detect AEs.  This has lead to the methods having high false positive rates \cite{Ryan2012, Reps2013a} as the majority of associations are not causal. Recent research has focused on using supervised techniques such as logistic regression \cite{Caster2013} to reduce the impact of confounding (i.e., when a hidden variable is responsible for the association). These supervised methods aim to distinguish between associations that are causal or non-causal by finding alternative causes of the medical event. Unfortunately, this requires generating a large number of regression models and also requires additional knowledge of possible confounders (e.g., other possible causes of the medical event).  Consequently, these methods are often slow and dependant on current knowledge.  Alternatively, a recent framework, side effect classifier (SEC), has been proposed that applies a single supervised classifier to identify ADRs efficiently \cite{Reps2014a} and the results suggest this framework is less susceptible to confounding.  

The SEC framework generates attributes inspired from the Bradford Hill causality criteria \cite{Hill1965}, a collection of factors that are often considered to determine causality, and uses these attributes and knowledge of existing ADRs to train a classifier capable of identifying new ADRs.  The framework was shown to identify ADRs with a low false positive rate and is highly efficient once the classifier is trained.  Different binary classifiers can be implemented within the framework, depending on the quantity of labelled data (this depends on the current knowledge of ADRs). It was noticed that the SEC framework implementing a different binary classifier will in general detect different ADRs, suggesting there is diversity between the classifiers.  Inspired by this diversity, in this paper we investigate whether using a type of ensemble, called a multiple classifier system, that combines predictions obtained from multiple classifiers is better than using a single classifier's prediction within the framework.  The multiple classifier system classifies each data-point corresponding to a drug-medical event pair as an ADR or non-ADRs based on a weighted combination of each individual classifiers confidence of the data-point belonging to the ADR class.  The weights are determined by using genetic algorithms to search for the values that optimise the SEC framework's ability to detect ADRs.   

This paper continues are follows. The next section gives an overview of genetic algorithms, multiple classifier systems and pharmacovigilance, including the SEC framework.  In section \ref{mat}, we described the longitudinal medical database used in this research, known as The Health Improvement Network (THIN) database (www.thin-uk.com), and highlight current issues with the data.  Section \ref{meth} describes the genetic algorithm method used to determine the classifier weights implemented by the multiple classifier system.  The results of the framework implementing the trained multiple classifier system is compared with the individual classifiers' results and are presented and discussed in section \ref{res}. The paper finishes with the conclusions in section \ref{conc}.

\section{Background}
\subsection{Genetic Algorithms}
Genetic algorithms are probabilistic search procedures inspired by the natural process of evolution \cite{Goldberg1988}.  The algorithm is an iterative process that initially starts with a population of candidate solutions that are randomly generated, and then these candidate solutions are evolved.  Each candidate solution has a set of genotypes (e.g., parameter values) and the set of genotypes determine the candidate solution's fitness.   During each iteration, a new generation of candidate solutions are created by recombination and mutation of the previous candidate solutions' genotypes based on their fitness.     

\subsection{Multiple Classifier System}
\begin{figure} \centering
\caption{The schema of a multiple classifier system.}
\label{mcs}
\includegraphics[width=0.45\textwidth]{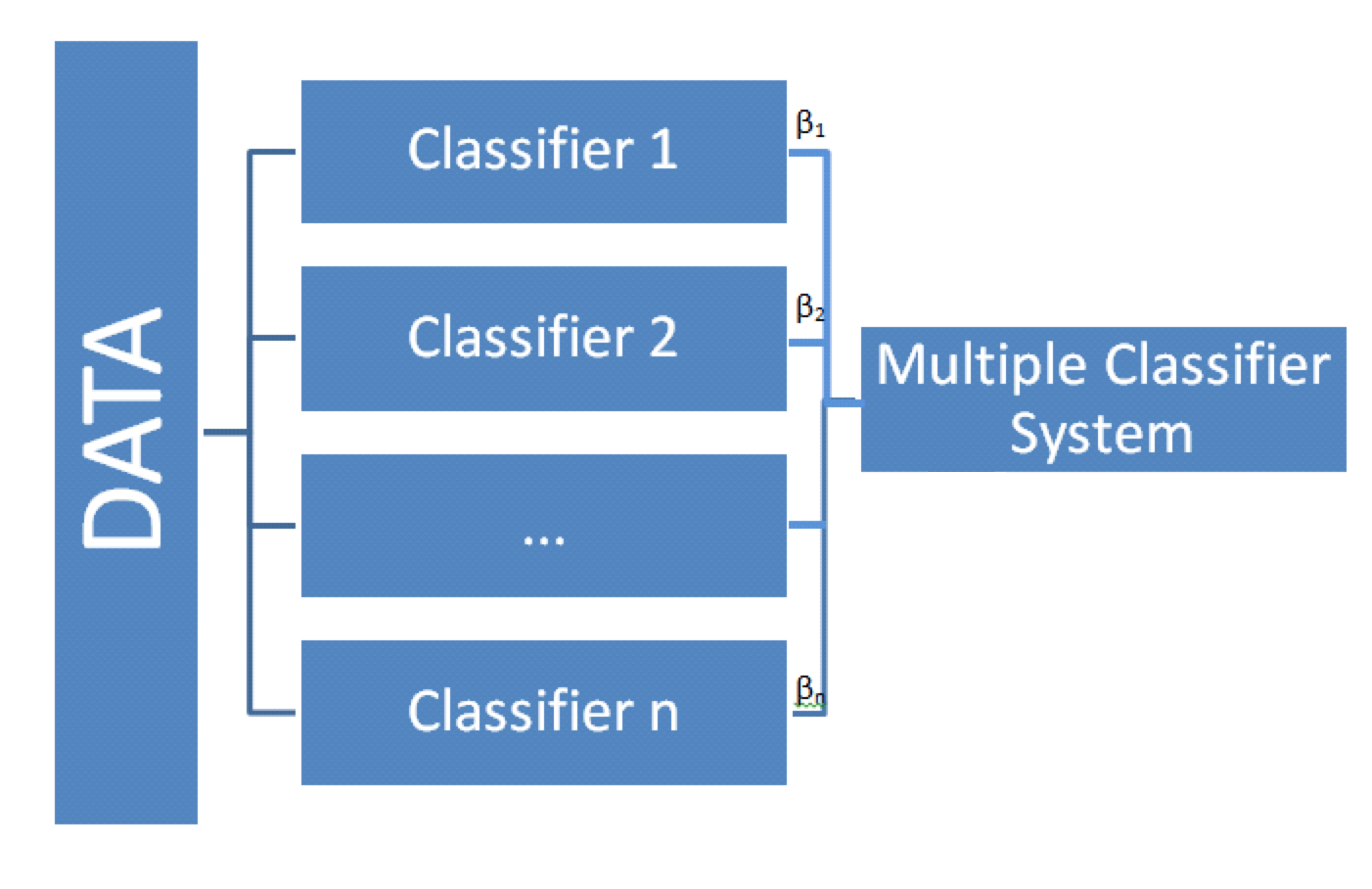}
\end{figure}

The term ensemble is used to describe a composition of multiple classifiers.  A type of ensemble that consists of a composition of various different classifiers has frequently been termed a multiple classifier system \cite{Windeatt2005} rather than called an ensemble.  This is to help distinguish between a combination of the same classifier trained with different perspectives (e.g., combining decision trees that are trained using different independent variables) and a combination of different classifiers (e.g., combining a SVM, a random forest, a neural network and a logistic regression model).  Fig. \ref{mcs} illustrates a multiple classifier system that combines the output of multiple single classifiers to generate a single output.  The aim of a multiple classifier system is the take advantage of diversity between classifiers to improve the classifying accuracy while maintaining efficiency.  Multiple classifier systems have been successfully implemented in numerous machine learning tasks including diagnosing melanoma \cite{Sboner2003}, classifying breast lesions \cite{Fusco2012} and detecting naked bodies in images \cite{Esposito2013}.  In the previous examples, combining multiple classifiers, under a suitable weighting scheme, was shown to improve performance compared to a single classifier.

As the classifiers used to identify ADRs within the SEC framework appear to be diverse, implementing a multiple classifier system that combines all the classifiers may improve the detection of ADRs.

\subsection{Previous Pharmacovigilance}
Pharmacovigilance is the study of prescription drug side effects. One important part of pharmacovigilance is the process of detecting drug side effects after the drugs have been approved and marketed.  Identifying drug side effects is a difficult task due to the majority of side effects relying on multiple factors, so it is common for some side effects to be observed rarely.  Clinical trails are unable to identify the majority of side effects prior to marketing  due to them only involving a small number of patients and being conducted under unrealistic conditions \cite{Corrigan2002}.  For example, patients involved in clinical trials are unlikely to take other drugs during the trial, so drug interactions can not be analysed.

\begin{figure} \centering
\caption{An example of an SRS database entity relationship diagram}
\label{srs2}
\includegraphics[width=0.5\textwidth]{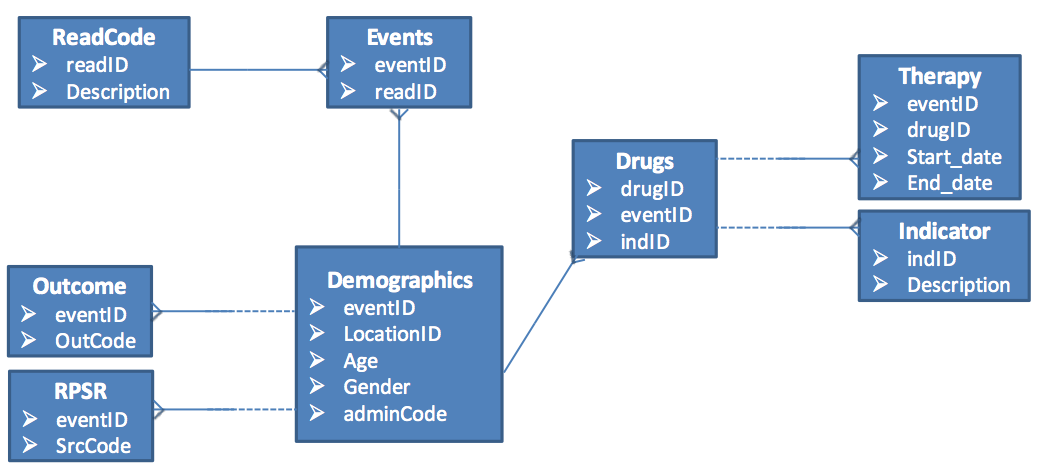}
\end{figure}

\begin{figure} \centering
\caption{Illustration of how the reports in the SRS database contain direct links between drugs and medical events.  Each report within the database consists of an observation of a patient taking a drug and then experiencing the medical event sometime after. }
\label{srs}
\includegraphics[width=0.5\textwidth]{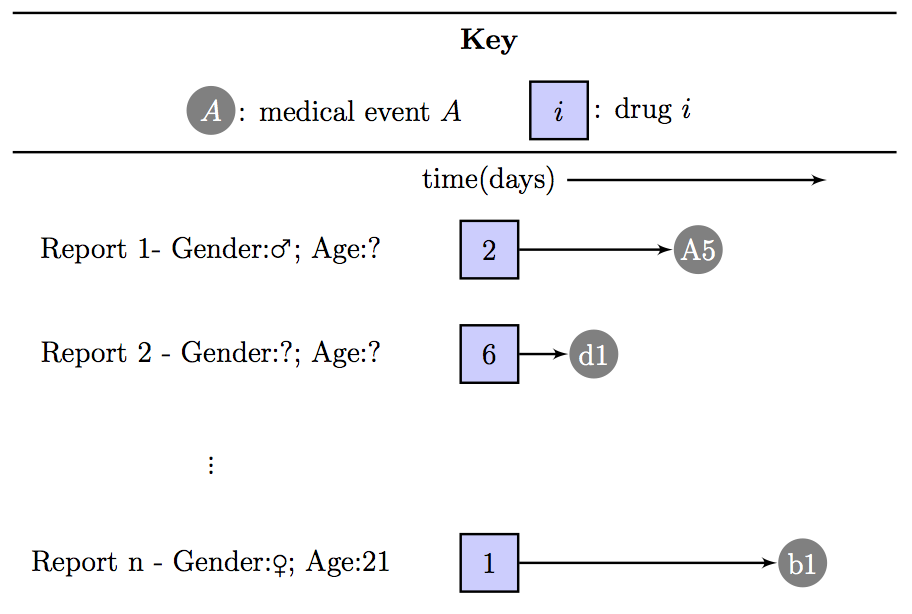}
\end{figure}

In general, the most widely implemented pharmacovigilance techniques have been developed for a specific type of medical database known as the spontaneous reporting system (SRS) databases \cite{Harmark2008}.  These databases consist of all the reports made by medical staff or the general public relating to a suspected ADR. The general design of the SRS databases is illustrated in Fig. \ref{srs2}.  The SRS databases contain natural links between drugs and medical events, see Fig. \ref{srs}. Sometimes additional information about the patient is included into the report, such as age and gender, but this is not compulsory. The techniques for detecting ADRs look for medical events linked disproportionally more to the drug than expected \cite{Van2002}.  Unfortunately, due to the reporting being voluntary, many ADRs may not be reported, and it is possible that some rare ADRs may never be noticed.  This under-reporting can prevent the early detection of ADRs and this means patients are put at risk for longer. In addition, there are known data quality issues such as missing, duplicated or incorrect data \cite{Lexchin2006}.

\begin{figure*} \centering
\caption{An example of inferring a link between a drug and medical event within a longitudinal healthcare database.  The medical events are represented by circles and the drugs represented by squares.  The potential acute ADRs are the medical events observed during the [$t_{0}$,$t_{1}$] time period centred around the prescription.}
\label{lhd}
\includegraphics[width=0.75\textwidth]{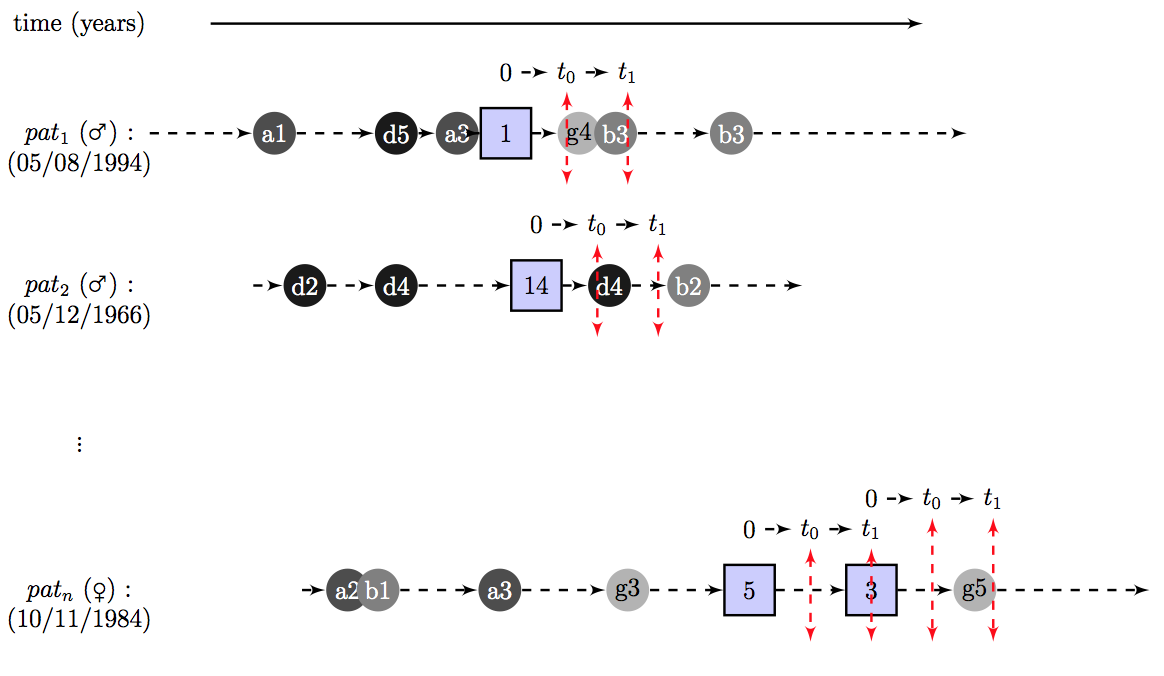}
\end{figure*}

Due to the limitations associated with the SRS databases, recent work has focused on using different types of medical databases \cite{Coloma2012}.  One example is the longitudinal healthcare databases.  These databases contain medical information about patients often spanning many years and it is common for them to contain records for millions of patients.  As this type of database does not rely on voluntary reporting, it presents a unique perspective for signalling ADRs.  However, it has been shown to suffer from different limitations.  The main limitation is that there are no clear links between drugs and medical events within the data itself, so potential links are inferred by finding the medical events that occur shortly after the drug in time.  This is illustrated in Fig. \ref{lhd}.  Unfortunately, the majority of the drug and medical events linked by time are associated but do no correspond to ADRs, and it has proved difficult for unsupervised algorithms to distinguish between the non-causal and causal relationships. 

\begin{figure} \centering
\caption{An example of a longitudinal healthcare database entity relationship diagram}
\label{lhd1}
\includegraphics[width=0.5\textwidth]{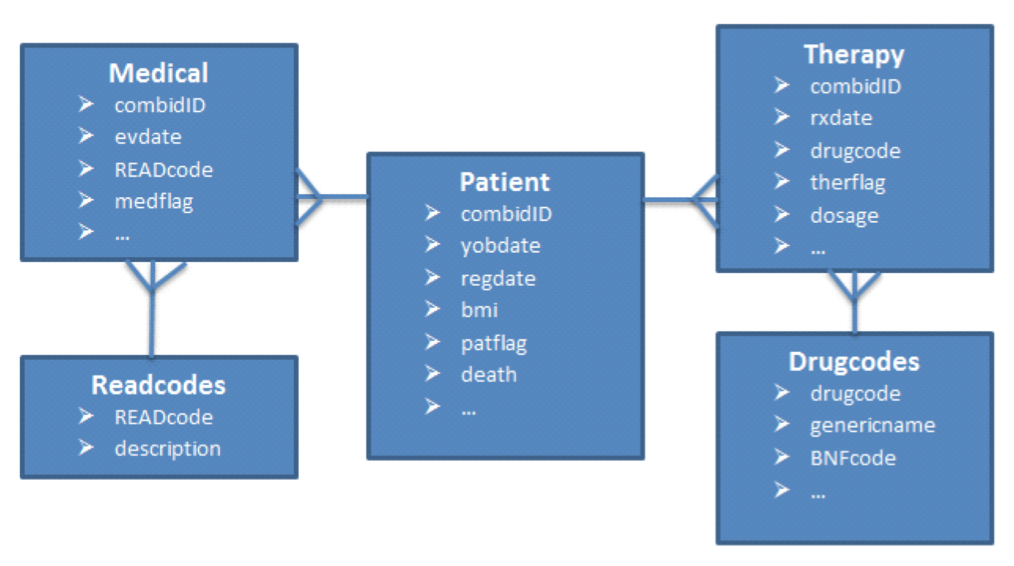}
\end{figure}

In \cite{Reps2012a}, the authors presented a semi-supervised algorithm that requires a user to input a drug of interest and then returns a ranked list of medical events. The higher a medical event is ranked by the algorithm, the more likely that medical event corresponds to a rare ADR of the specified drug of interest.  The algorithm generated the data by extracting attributes that are insightful for ADR detection from a longitudinal healthcare database and determined labels for some medical events by mining online medical websites.  The labelled and unlabelled data were then used to clusters similar medical events into either an ADR cluster, an indicator (a cause of taking the drug) cluster or a noise cluster.  Medical events assigned to the noise cluster were filtered, and the remaining medical events where ranked based on how often they occurred after the drug divided by how often they occurred before the drug multiplied by a cluster dependent weight. 

The success of the semi-supervised algorithm them prompted the idea of generating causal inference based attributes for a selection of drug-medical event pairs that are definitively known ADRs or non-ADRs \cite{Reps2012c} and using this data to train a classifier that can then be used to predict new ADRs.  One such supervised framework generated attributes based on the counterfactual theory of causality \cite{Reps2013b}, whereas another framework, SEC, generated attributes based on the Bradford Hill causality criteria. Rather than mining online forums for the known ADR and non-ADR labels, both frameworks used an online resource that contains lists of ADRs that were mined from drug packaging.  Both supervised frameworks demonstrated excellent performance and previous results suggest supervised techniques may help improve current pharmacovigilance.  

\subsubsection{SEC Framework}
The previously presented SEC framework is a supervised algorithm for detecting ADRs. The algorithm automates the technique of inferring causality via the Bradford Hill causality criteria, as this technique is commonly applied to assess whether a side effect is caused by a drug or not.  The SEC framework requires three steps.  The first step is data generation where suitable labelled data are extracted for each drug-medical event pair that represent a possible acute ADR.  The second step is training a binary classifier using the labelled data to classify each drug-medical event pair as an ADR or non-ADR, and the final step is applying the trained classifier to new unlabelled data.\\

\textbf{Step 1) Data generation}\\
As we are interesting in detecting acutely occurring ADRs, we find the drug-medical event pairs that are possible ADRs by investigating the medical events that occur within a month of a drug being prescribed.  To train a binary classifier we need a set of attribute vectors $\mathbf{x_{i}} \in \mathbb{R}^{n}$ and their corresponding class $y_{i} \in \{-1,1\}$.  In the SEC framework, each data point corresponds to a drug-medical event pair of interest, where the i\textsuperscript{th} drug-medical event pair has the attribute vector $\mathbf{x_{i}}$ and class $y_{i}$.  Therefore, to generate the training data, the first step is to identify the drug-medical event pairs of interest, the second step is to determine their labels  and the final step is to calculate their attributes.

To identify the drug-medical events pairs of interest, we restrict out attention to a set of specified drugs, denoted by $D$.  For each drug $d_{i} \in D$, we use temporal relationships to identify the risk medical events of $d_{i}$ ($RME_{d_{i}}$).  The risk medical events of $d_{i}$ are the medical events that were observed during the month after a prescription of $d_{i}$ for one or more patients, $RME_{d_{i}}=\{$medical events $|$ the medical event occurs within a month of $d_{i} $ for one or more patients $\}$.  The drug-medical event pairs of interest are all the possible combinations of $d$-$e$, where $d \in D$ and $e \in RME_{d}$.  The drug-medical event pairs of interested with labels are then determined. For the i\textsuperscript{th} drug-medical event pair, if the medical event is labelled as a known side effect of the drug within the online drug resource known as SIDER \cite{Kuhn2010}, then the pair is labelled as an ADR ($y_{i}=1$).  Alternatively, if the medical event cannot possibly correspond to an acute ADR (e.g, the medical event is `cancer', `menopause' or `death of family member'), the drug-medical event is labelled as a non-ADR ($y_{i}=-1$).  Any drug-medical event pair neither listed on SIDER as corresponding to a known ADR nor clearly a non-ADR is ignored as the pair has no definitive label.

For the $i$\textsuperscript{th} drug-medical event pair labelled as an ADR or non-ADR, we calculate the Bradford Hill causality criteria based attributes, described in \cite{Reps2014a} and denote the vector consisting of these attributes by $\mathbf{x_{i}}$.  The attributes are derived from a selection of the Bradford Hill causality criteria:
\begin{itemize}
\item \textbf{Association strength:} How strong the association between the drug and medical event is.
\item \textbf{Temporality:} Does the drug precede the medical event or the other way?
\item \textbf{Specificity:} How specific the medical event  is, or how similar patients experiencing the medical event are.  
item \textbf{Biological gradient:} Measures whether the probability of the medical event increases as the drug dosage increases. 
\item \textbf{Experimentation:} Does the medical event start and stop when the drug starts and stops?
\end{itemize}

In summary, for the $i$\textsuperscript{th} labelled drug-medical event pair we have $(\mathbf{x_{i}}, y_{i})$, where $\mathbf{x_{i}}$ is the Bradford Hill causality attributes and $y_{i}=1$ when the $i$\textsuperscript{th} drug-medical event pair is a known ADR and $y_{i}=-1$ when the $i$\textsuperscript{th} drug-medical event pair is a known non-ADR.  The complete set of labelled data is denoted by $X$, where $X=\{ (\mathbf{x_{i}},y_{i})  \}$.\\

\textbf{Step 2)Training a binary classifier}\\
The labelled data are then used to train a binary classifier (the choice of classifier is determined by the user as any classifier can be used within the framework), 
\begin{equation}
f:X \to Y; f(\mathbf{x_{i}}) \to \{-1,1 \}
\end{equation}
where $f(\mathbf{x_{i}})=-1$ means the drug-medical event pair is classified as a non-ADR and $f(\mathbf{x_{i}})=1$ means the drug-medical event pair is classified as an ADR.  The chosen classifier is trained using ten-fold cross validation to reduce overfitting.  In previous work \cite{Reps2014a}, the random forest classifier was found to perform better than a support vector machine, a logistic regression and a naive Bayes classifier.  \\

\textbf{Step 3)Applying trained classifier}\\
The trained classifier is then applied to the attribute vector $\mathbf{x_{*}}$ for a new drug-medical event pair, and the prediction $f(\mathbf{x_{*}})$ is returned.  

For evaluating the framework, the labelled data are partitioned into training/testing data and validation data.  The training/testing data are used to train the classifier and the validation data are used to evaluate the performance of the trained classifier by comparing the predicted class with the true class. 

\section{Materials} \label{mat}
\begin{figure} \centering
\caption{The patient table within the THIN database.}
\label{pat}
\includegraphics[width=0.5\textwidth]{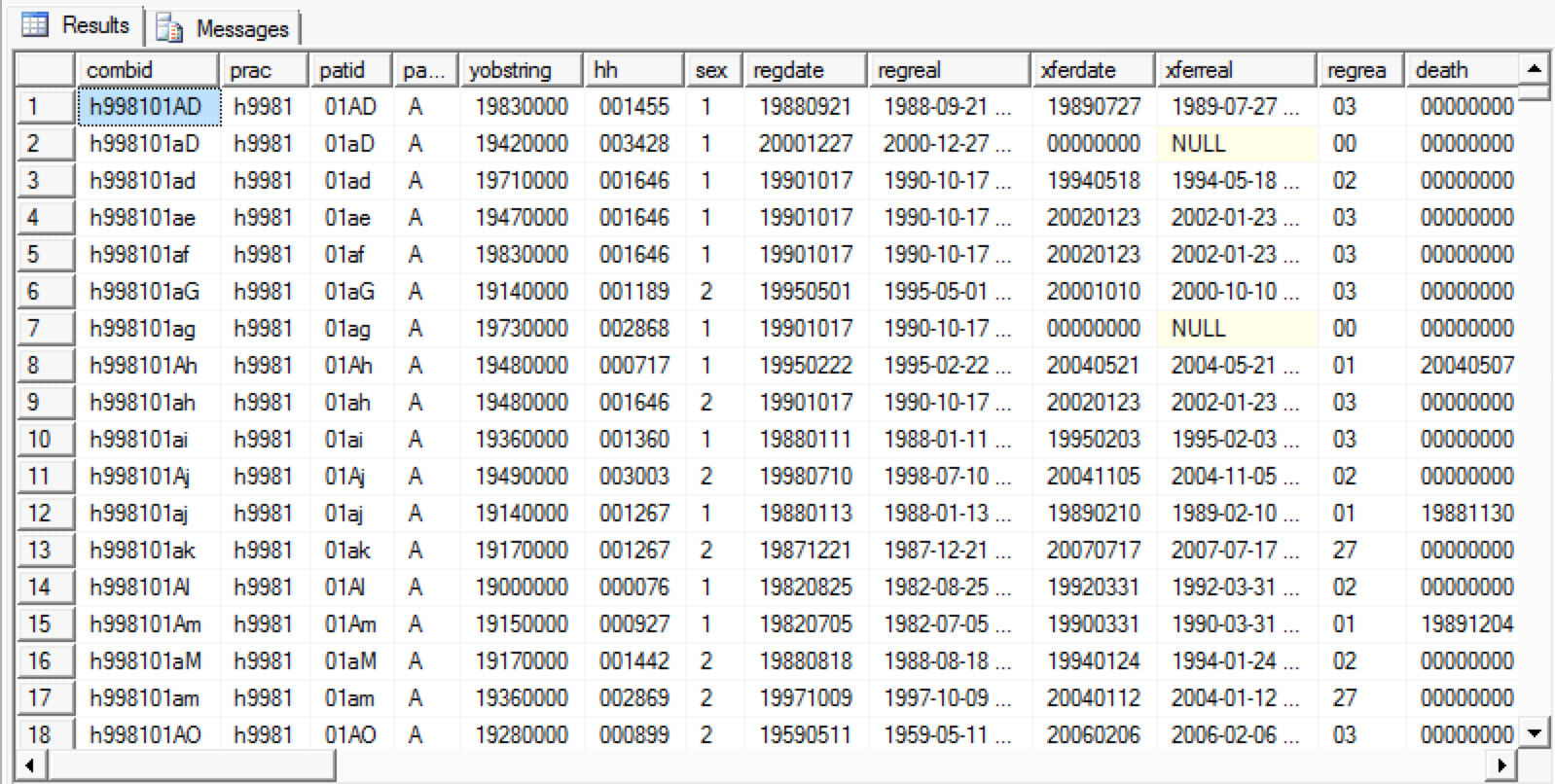}
\end{figure}

\begin{figure} \centering
\caption{The medical table within the THIN database.}
\label{med}
\includegraphics[width=0.5\textwidth]{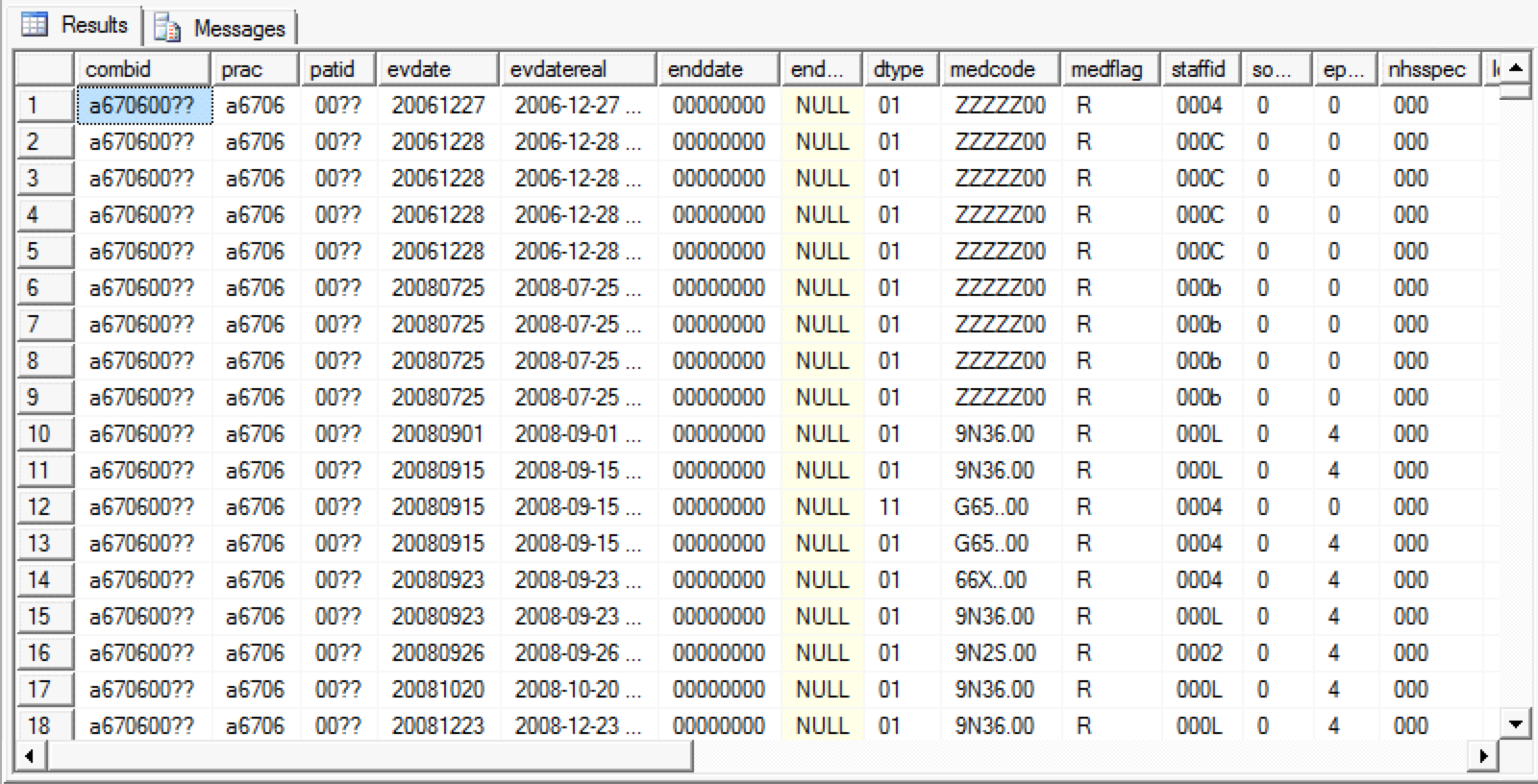}
\end{figure}

\begin{figure} \centering
\caption{The therapy table within the THIN database.}
\label{ther}
\includegraphics[width=0.5\textwidth]{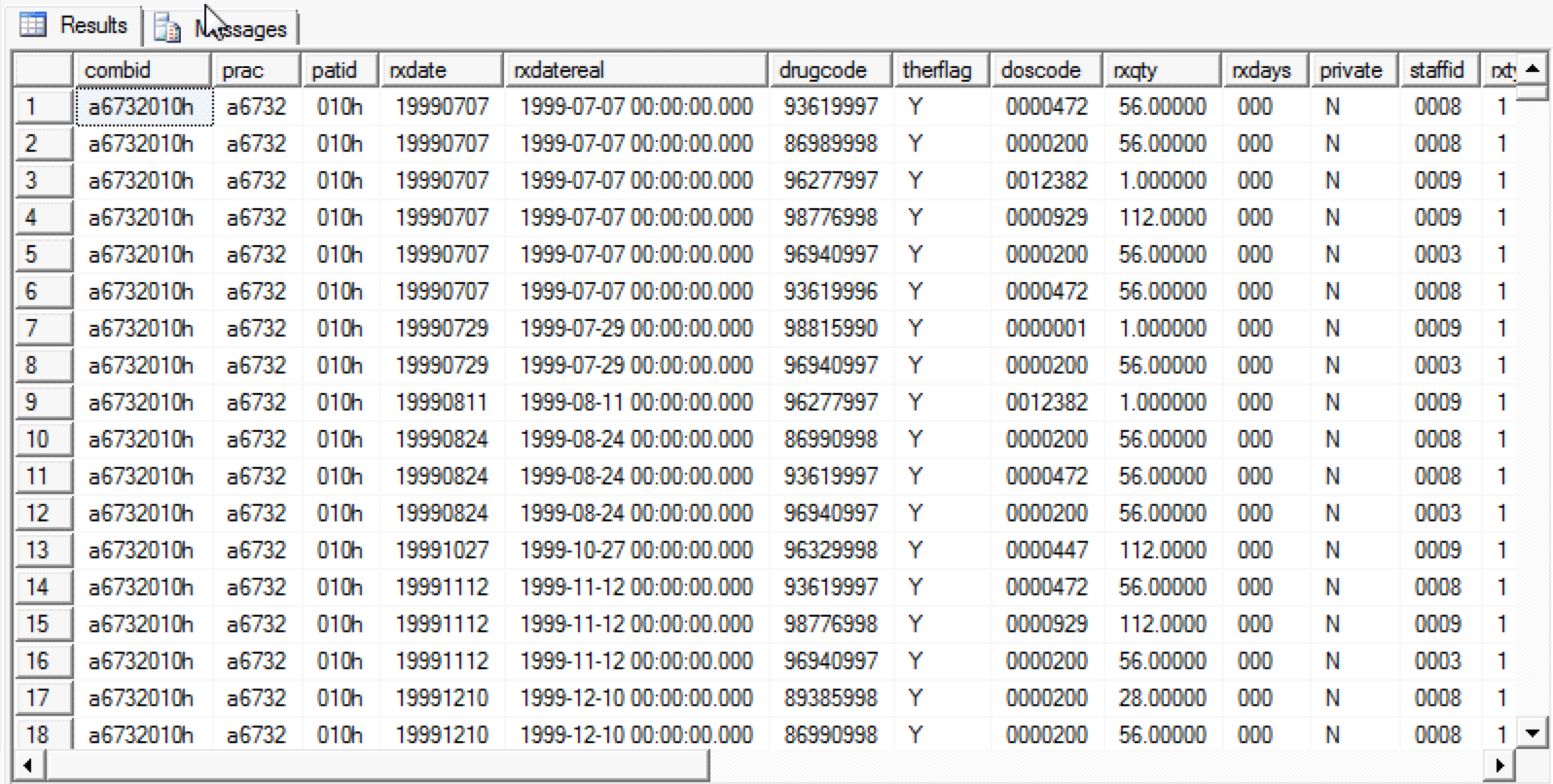}
\end{figure}

The THIN database contains temporal medical data for over 11 million patients (approximately 4 million currently active patients).  The data is anonymised, so each patient is represented by a unique patient ID rather than the patients real name.  There are three main tables within the THIN database, the patient table, the medical table and the therapy table, see Figs. \ref{pat}-\ref{ther}.  The patient table contains personal information about each patient in the database including their year of birth, their gender and their date of registration.  The therapy table contains timestamped records of each patient's drug prescription history, so each record includes the patient ID, the date of the prescription and information about the prescription (drug details and dose details).  The medical table is similar to the therapy table but contains timestamped records of each patient's medical event history (i.e., illnesses, diseases, laboratory tests and administrative events), so a typical record contains the patient ID, the date of the medical event and the medical event information, recorded via the READ codes.

Each READ codes consist of five elements from the alphabet $\{a-z,A-Z,1-9, \dot  \}$ and they have a hierarchal structure.  The depth of a node within a tree is the length of the minimum path from the node to the root.  Unfortunately, the READ codes have redundancies and the same medical event can be represented by various distinct READ codes.  This can cause issues for data miners, however the SEC algorithm generates attributes specifically to prevent this issue having a negative effect on its ability to detect ADR.

\section{Methodology} \label{meth}
In this paper we are developing a multiple classifier system to be implemented within the SEC framework and comparing its ability to detect side effects with the framework implementing a single classifier.  Therefore, in this section the methods used to analyse the single classifier framework and the multiple classifier system framework are both described.

To evaluate each framework, we determine all the labelled drug-medical event pairs correspond to the $25$ drugs: nifedipine, amlodipine, felodipine, nicardipine, verapamil, ciprofloxacin, ofoxacin, norfloxacin, nalidixic acid, moxifloxacin, fluconazole, itraconazole, posaconazole, voriconazole, ibuprofen, fenoprofen, ketoprofen, celecoxib, flurbiprofen, nabumetone, naproxen, budesonide, beclometasone, hydrocortisone and prednisolone.  These labelled data are composed of the $30$ Bradford Hill causality criteria derived attributes for each drug-medical event data-point and a label specifying whether the drug-medical event data-point is listed as an ADR on SIDER or one of the manually selected non-ADRs.

There were a total of $5710$ drug-medical event data points with known labels corresponding to the $25$ chosen drugs.  The labelled data were partitioned into training/testing data $X_{T}$ (80\% of the labelled data) and validation data $X_{V}$ (20\% of the labelled data).  The training/testing date were used to train the classifier or multiple classifier system and the validation data were used to evaluate the framework implementing the single classifier or multiple classifier system.

The measure used to determine the effectiveness of each framework is the area under the receiver operating characteristic curve.  This measure corresponds to the probability of a drug-medical event pair known to be an ADR being assigned a higher confidence of being within the ADR class by the framework than a drug-medical event pair known to be a non-ADR \cite{Bradley1997}.  In particular, we restrict out attention to a partial area, as we are only interested in the section of the curve where few drug-medical event pairs are classed as side effects \cite{Jiang1996}.  When many drug-medical event pairs are classed as ADRs, there are likely to be many non-ADRs pairs incorrectly classed as ADRs and this is undesirable.  The partial area under the curve that we are interested in is denoted by pAUC$_{[0.9,1]}$ and a more detailed explanation of how the measure is calculated can be found in section \ref{ev}.

\subsection{SEC Framework: Single Classifier}
To analyse the single classifier framework, the SEC framework implementing either a random forest, support vector machine, logistic regression, naive Bayes or k-nearest neighbours classifier is trained using ten-fold cross validation on the training/testing data $X_{T}$.  The trained classifier is denoted by $f: \mathbb{R}^{30} \to \{-1,1\}$, where $f(\mathbf{x_{i}})=-1$ represents the $i$\textsuperscript{th} drug-medical event pair being classifier as a non-ADR and $f(\mathbf{x_{i}})=1$ represents the $i$\textsuperscript{th} drug-medical event pair being classifier as an ADR.

\subsection{SEC Framework: Ensemble Classifier}
The multiple classifier system framework requires training multiple classifiers and learning the optimal weighted combination of the classifiers. In this framework, after the training data is generated, the data is firstly used to train various classifiers and then used to determine a weighted combination of all the classifier.

\subsubsection{Training the classifiers}
Five classifiers (random forest, support vector machine, logistic regression, naive Bayes and k-nearest neighbours) are trained via ten fold cross validation to determine the optimal parameters that maximise the partial area of interest under the curve (pAUC$_{[0.9,1]}$, see section \ref{ev}) using the training/testing $X_{T}$ set.  Each classifier is trained using a grid search over suitable parameter values, these can be seen in Table \ref{tab_classifiers} and the chosen parameter values are also listed.
\begin{table*} \centering
\caption{The different classifiers used by the multiple classifier system and their optimal parameters.}
\label{tab_classifiers}
\begin{tabular}{lll}
Classifier & Parameters :-(grid search range) & Optimal Parameters \\ \hline \hline
$f_{1}$: Random Forest & mtry:-[1,30] & mtry=11 \\
$f_{2}$: Support Vector Machine (Radial)& sigma:-(0,1], C:-(0,10] & sigma=0.0978, C=6.1624 \\
$f_{3}$: K-Nearest Neighbours & K:-[1,100] & K= 17\\
$f_{4}$: Logistic Regression & decay:-[0,10] & decay=0 \\
$f_{5}$: Naive Bayes & fL:-[0,1], usekernel:-$\{$ TRUE,FALSE $\}$ & fL=0, usekernel=TRUE \\  
\end{tabular}
\end{table*}

For each trained classifier $f_{i}$, we can also extract the classifiers confidence that the drug-medical event is in the ADR class, this is denoted by $c_{i}:\mathbb{R}^{30} \to [0,1]$.  So $c_{i}(\mathbf{x_{j}})$ is the confidence of the i \textsuperscript{th} classifier that the j \textsuperscript{th} drug-medical event pair is an ADR.  

\subsubsection{Determining the weights}
Using these confidence functions, genetic algorithms are applied to find the optimal weights $\beta_{i}, i \in [1,5]$ for the multiple classifier system that determines the class of the $j$\textsuperscript{th} drug-medical event pair by,
\begin{equation}
f_{6}(\mathbf{x_{j}}) =\left \{ \begin{array}{ll}
1 & if \sum_{i} \beta_{i} c_{i}(\mathbf{x_{j}}) \geq \alpha \in (0,1) \\
 -1 & \mbox{ otherwise}\\ \end{array} \right.
\end{equation}
The value $\alpha$ is the natural threshold and this controls the stringency of the multiple classifier system.

\begin{table*} \centering
\caption{The genetic algorithm parameters.}
\label{ga}
\begin{tabular}{p{0.07\textwidth}p{0.13\textwidth}p{0.14\textwidth}p{0.05\textwidth}p{0.25\textwidth}p{0.10\textwidth}p{0.08\textwidth}}
Population size & Crossover type & Mutation type & Elitism used & Selection criteria & Initialisation & Stopping criteria \\ \hline \hline
1000 & Local arithmetic crossover & Uniform random mutation & True & Fitness proportional selection with fitness linear scaling & Uniformly chosen from [0,1] & After 500 iterations \\

\end{tabular}
\end{table*}

The weights are determined by implementing a genetic algorithm with a mutation rate of 0.1 and applying elitism with a candidate population size of 1000 until convergence, see Table \ref{ga} for full details.  The fitness of each weight vector ($\boldsymbol{\beta}$) is the ten fold cross validation average of the the partial AUC over the specificity range [0.9,1] for the multiple classifier system based on that weight scheme on the training/testing set.  The optimal weight vector was,
\begin{equation} \begin{split}
\boldsymbol{\beta}&= (\beta_{1}, \beta_{2}, \beta_{3}, \beta_{4}, \beta_{5})\\
                             &=(0.701, 0.314, 0.002, 0.026, 0.012)
\end{split}
\end{equation}
where $c_{1}()$ is random forest,  $c_{2}()$ is support vector machine, $c_{3}()$ is K-nearest neighbours, $c_{4}()$ is logistic regression and $c_{5}()$ is naive Bayes.

\subsection{Evaluation} \label{ev}
The framework implementing a single trained classifier or the multiple classifier system is then applied to the validation set and the prediction of each data-point in the validation set is compared with the truth. The number of true positives (TP), false positives (FP), false negatives (FN) and true negatives (TN) are calculated as follows,
\begin{description}
\item[\textbf{TP}:] $|\{i | y_{i}=f(\mathbf{x_{i}})=1  \}|$
\item[\textbf{FP}:] $|\{i | y_{i}=-1,f(\mathbf{x_{i}})=1  \}|$
\item[\textbf{FN}:] $|\{i | y_{i}=1,f(\mathbf{x_{i}})=-1  \}|$
\item[\textbf{TN}:] $|\{i | y_{i}=f(\mathbf{x_{i}})=-1  \}|$
\end{description}
Using the above values, the accuracy, precision, sensitivity, and specificity can be calculated, 
\begin{equation} 
\begin{split}
\mbox{Sensitivity} &= \mbox{(TP)}/\mbox{(TP+FN)} \\
\mbox{Specificity} &= \mbox{(TN)}/\mbox{(TN+FP)} \\
\mbox{Accuracy} &= \mbox{(TP+TN)}/\mbox{(TP+FP+FN+TN)} \\
\mbox{Precision} &= \mbox{(TP)}/\mbox{(TP+FP)} 
\end{split}
\end{equation}
The receiver operating characteristic (ROC) curve is generated by potting the sensitivity against 1 minus the specificity and the AUC is the area under this curve.  The AUC measures the general ability of a classifier rather than only considering how well it does it at its natural threshold and is a fairer measure for comparing different classifiers. The pAUC$_{[0.9,1]}$ is the partial area under the ROC curve, between the specificity values of $0.9-1$, this value is useful as we are interested in the classifiers ability when the specificity is high and the number of of false positives is low.

\section{Results \& Discussion} \label{res}
\begin{table*} \centering
\caption{The results of the SEC framework implementing a single classifier or multiple classifier system for the validation set.}
\label{results}
\begin{tabular}{llllll}
Framework classifier & Accuracy & Precision & Sensitivity & Specificity & pAUC$_[0.9,1]$ \\ \hline \hline
$f_{1}$: Random Forest & 0.930 & 0.789 & 0.380  & 0.989 & 0.769 \\
$f_{2}$: Support Vector Machine & 0.921 & \textbf{0.809} & 0.241  & \textbf{0.994} &0.710 \\
$f_{3}$: K-Nearest neighbours & 0.917 & 0.729 & 0.222  & 0.991 & 0.695 \\
$f_{4}$: Logistic Regression & 0.086 & 0.085 & \textbf{0.861}& 0.003 &0.693 \\
$f_{5}$: Naive Bayes & 0.912 & 0.577 & 0.354 & 0.972 & 0.710 \\
$f_{6}$: Multiple Classifier System & \textbf{0.933} & 0.782 & 0.430  & 0.987 & \textbf{0.772} \\
\end{tabular}
\end{table*}

\begin{figure}\centering
\caption{The ROC plots for the frameworks ability to detect ADRs when implementing the different classifiers.}
\label{roc}
\includegraphics[trim=0.4cm 0cm 0cm 0cm, clip=true,width=0.5\textwidth]{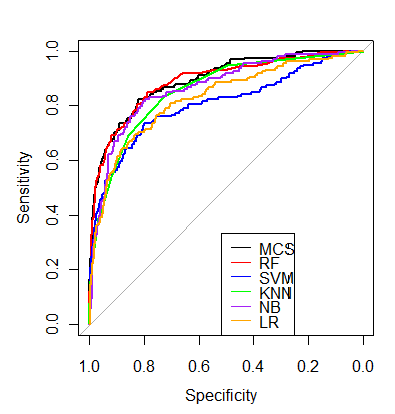}
\end{figure}

The results are presented in Table \ref{results} and ROC plots for the framework implementing the range of classifiers or the multiple classifier system can be seen in Fig. \ref{roc}.  The optimal value for $\alpha$ (the multiple classifier system's natural threshold) was found to be $\alpha=0.4381$.  It can be seen that the framework implementing a multiple classifier system ($f_{6}$) obtained a superior accuracy,  sensitivity  and pAUC$_{[0.9,1]}$ than the framework implementing any single classifier.  However, using a bootstrap test to compare the pAUC$_{[0.9,1]}$s \cite{Pepe2008} of the random forest and the multiple classifier system at a 5\% significance level, the pAUC$_{[0.9,1]}$ was not shown to be significantly different (p-value=0.499).  The highest precision and specificity values were obtained by the framework implementing a support vector machine and not the multiple classifier system. This is probably due to the multiple classifier system being optimised specifically for the partial AUC. If the precision or specificity was deemed to be more important, different weights could be calculated by the genetic algorithm to optimise the multiple classifier system for the desired measure (e.g., precision or specificity).  

The ensemble weights do not necessarily reflect the importance of the classifier within the ensemble, as each classifier has varying ranges for its confidence function values. It may be useful to normalise the confidence function values prior to determining the optimal ensemble weights. If the classifier confidence weights were normalised, then the ensemble weights would correspond to the importance of the classifier and this would then help indicate which of the classifiers was most influential within the ensemble.  This knowledge could be used to remove classifiers that had little influence.

The advantage of ensemble approaches rather than relying on any individual classifier is that they generally reduce the classifier's variance. This is useful for ADR detection, as the training set is likely to change and grow as new ADRs are discovered. An ensemble approach for ADR detection is also useful, as previous results have shown that each classifier tends to make different mistakes, so the ensemble can overcome an individual classifiers misclassification. This is the likely reason why the ensemble obtained an improved performance. However, the disadvantages are that the ensemble is computationally longer due to the requirement of training multiple classifiers and then tuning the ensemble weights. Although the multiple classifier system improved the accuracy and $pAUC_{[0.9,1]}$ compared to each single classifier, the improvement was not significant. This may suggest that when the training data is sufficiently large to enable good performance from a single classifier, the small benefit in performance of the ensemble is not enough to overcome the extra cost of complexity. It would be interesting to investigate how the ensemble performs relative to each individual classifier at various training set sizes.

\section{Conclusions}\label{conc}

In previous work, it was shown that different classifiers detected different side effects. In this paper we combined various classifiers with the aim of improving the overall discovery of side effects.  The classifiers were combined using genetic algorithms to tune a multiple classifier system that can be used within a side effect discovery framework. We then compared the side effect discovery framework implementing a multiple classifier system with the framework implementing a single classifier. The results show that a larger partial AUC can be obtained by a multiple classifier system that integrates multiple diverse classifier by calculating a weighted aggregate of their confidences that a data-point belongs to the class ADR.  This research presents a novel useful application of genetic algorithms.

Possible areas of future work could investigate using a suitable evolutionary algorithm to tune each of the individual classifiers rather than using a grid search (i.e., a selection of values for each parameter in input and the search is done over all possible parameter combinations), as this may increase their individual performance in addition to the multiple system classifiers performance.



\bibliographystyle{IEEEtran}
\bibliography{ev_ensemble_ref}
%

\def\V{\rm vol.~}
\def\N{no.~}
\def\pp{pp.~}
\def\Pot{\it Proc. }
\def\IJCNN{\it International Joint Conference on Neural Networks\rm }
\def\ACC{\it American Control Conference\rm }
\def\SMC{\it IEEE Trans. Systems\rm , \it Man\rm , and \it Cybernetics\rm }

\def\handb{ \it Handbook of Intelligent Control: Neural\rm , \it
    Fuzzy\rm , \it and Adaptive Approaches \rm }

\end{document}